\documentclass[conference]{IEEEtran}
\usepackage{cite}
\usepackage{amsmath,amssymb,amsfonts}
\usepackage{algorithmic}
\usepackage{graphicx}
\usepackage{textcomp}
\usepackage{xcolor}
\usepackage{multirow}
\usepackage{amsmath}
\def\BibTeX{{\rm B\kern-.05em{\sc i\kern-.025em b}\kern-.08em
    T\kern-.1667em\lower.7ex\hbox{E}\kern-.125emX}}
\begin{document}

\title{Vision-Based Approach for Food Weight Estimation
from 2D Images\\
\thanks{Identify applicable funding agency here. If none, delete this.}
}

\author{\IEEEauthorblockN{Chathura Wimalasiri}
\IEEEauthorblockA{\textit{Dept. of Computer Engineering} \\
\textit{University of Peradeniya}\\
Peradeniya, Sri Lanka \\
e18402@eng.pdn.ac.lk }
\and
\IEEEauthorblockN{Prasan Kumar Sahoo}
\IEEEauthorblockA{\textit{Dept. of Computer Science and
Information Engineering} \\
\textit{Chang
Gung University}\\
Guishan, Taiwan \\
pksahoo@mail.cgu.edu.tw}
}

\maketitle

\begin{abstract}
In response to the increasing demand for efficient and non-invasive methods to estimate food weight, this paper presents a vision-based approach utilizing 2D images. The study employs a dataset of 2380 images comprising fourteen different food types in various portions, orientations, and containers. The proposed methodology integrates deep learning and computer vision techniques, specifically employing Faster R-CNN for food detection and MobileNetV3 for weight estimation. The detection model achieved a mean average precision (mAP) of 83.41\%, an average Intersection over Union (IoU) of 91.82\%, and a classification accuracy of 100\%. For weight estimation, the model demonstrated a root mean squared error (RMSE) of 6.3204, a mean absolute percentage error (MAPE) of 0.0640\%, and an R-squared value of 98.65\%. The study underscores the potential applications of this technology in healthcare for nutrition counseling, fitness and wellness for dietary intake assessment, and smart food storage solutions to reduce waste. The results indicate that the combination of Faster R-CNN and MobileNetV3 provides a robust framework for accurate food weight estimation from 2D images, showcasing the synergy of computer vision and deep learning in practical applications.
\end{abstract}


\section{Introduction}
In today's digital world, food is more important than just a source of strength and nutrients. With the increasing use of technology, 2D food images are appearing everywhere and influencing our food choices. There are several uses of 2D food images that can be used to visualize and analyze data, food recognition and classification, monitor food quality and defect detection, assess food safety and hygiene practices, and create visually appealing and appetizing representations of food. With these facts, human life is getting easier and safer.

Artificial intelligence has emerged as a trans-formative force in the modern world, impacting virtually every aspect of our lives. Deep learning is a subset of artificial intelligence. Deep learning can perform complex tasks such as image recognition, natural language processing, and decision-making. A convolutional neural network (CNN) is a common deep learning architecture. CNNs recognize patterns and extract features from grid-like data, making them the go-to solution for various computer vision tasks. Computer vision is a sub-field of AI that is used for classification, object detection, and segmentation. Combining both computer vision and deep learning concepts can be done with the most useful real-world applications, such as self-driving cars, medical image analysis, facial recognition, object detection and tracking, and image and video captioning. 

Food weight estimation using artificial intelligence is an important step in real life. This concept can be applied to automatically estimate calorie intake based on pictures of meals by estimating food weight.  Smart food storage containers can use this concept to reduce food waste, store them in proper conditions, maximize their shelf life, and reduce spoilage. Restaurants can use this to accurately measure the food weight using 2D images. This can be used in medical applications to monitor food and calorie amounts easily.

In this paper, we propose a method to estimate food weight from 2D images using deep learning and computer vision techniques. We use a novel dataset, which includes 2380 images with fourteen types of food in 2D with different portions, orientations, and containers. Food types and information are presented in Table~\ref{tab1}. After a food image is acquired, the first challenge is to recognize it. After that, using these results, weight estimation is performed. 

The paper is structured as follows: In the next section, a brief overview of the related works in this area will be provided. Next, we describe our food recognition and weight estimation methods. Next, we discuss our results in food recognition and weight estimation. Finally, we summarize our work.

\section{Related Works}
This paper is mainly focused on estimating food weight from 2D images using a vision-based approach. This aligns with various food-related tasks, including calorie estimation. Research by\cite{b1} proposed a method to find calories in Indonesian street food using 2D images. Before finding the calorie amount\cite{b1}, find the weight of the food item using multiple linear regression. Then,\cite{b1} used a weight-calorie scale to find the amount of calories in food. 

Research by\cite{b2} proposed a method to find the weight of lettuce using a 3D stereoscopic technique.\cite{b2} proposed a new image preprocessing technique to estimate weight using a 3D stereoscopic technique and estimated the weight of fresh lettuce from a 3D spatial domain. Research by \cite{b3} proposed a method to localize the picking point of strawberries and estimate their weight.\cite{b3} proposed two novel datasets annotated with picking points, key points, weight, and size of strawberries.\cite{b3} used the concept that the weight of a strawberry is proportional to its shape, size, and density in their weight estimation. Research by \cite{b4} proposed a method to estimate the weight of meals in self-service lunch line restaurants using 2D images.\cite{b4} capture the top view of the meal, identify each food item, and estimate the weight of each item on the plate. Research by\cite{b5} proposed a method to estimate the weight of harumanis mango from an RGB image. All the mango image dataset background is an A4 sheet to reduce background noise and to function as a visual cue for CNNs. \cite{b5} defined a CNN architecture and the final layer is a dense layer with rectified linear unit (ReLU) activation function for regression task. Research by \cite{b6} proposed a method to estimate the weight of melons from unmanned aerial vehicle images. \cite{b6} detected all melons in an image and estimated their weight individually, finally estimating the weight of the melons yield. Research by \cite{b7} proposed a smart harvesting decision system to estimate fruits types,  maturity level, and weight. \cite{b7} used a powerful supervised machine learning algorithm called support vector machine (SVM) to estimate weight. Weight isn't a discrete number; therefore, \cite{b7} used a regression technique called support vector regression (SVR).  Research by \cite{b8} proposed a method to analyze food calories and nutrition. For that task, \cite{b8} estimated the weight of food using a linear regression model and, from that, estimated food calories and nutrition. Research by \cite{b9} proposed a method of dietary assessment for Chinese children. \cite{b9} estimated the weight of food using 2D images in the system in order to achieve their goal. They stored density information in their database and calculated the volume and estimated the weight using the relationship between weight, volume, and density. Research by \cite{b10} proposed a method to estimate food portions from a single view using geometric models. 3D reconstruction using a single view is a very challenging task; therefore, they have used geometrical models such as the shape of the container. \cite{b10} estimated the volume of each food with the help of the geometric models, then relationships between weight, volume, and density were used to estimate the weight of the food item. Research by \cite{b11} proposed a method to estimate food intake using a depth camera. First, the empty tray, before eating and after eating were captured, respectively. Using these three image information, the volume of food intake was estimated. After that, the weight was estimated using a specific gravity function. Finally, weight was converted to calories using food calorie database information. Research by \cite{b12} proposed a dietary assessment method to record daily food intake using a food image of a meal. First, the system identified and segmented food in the image. Then, they used a 3D reconstruction method for regular-shaped foods and an area-based weight estimation method for irregular-shaped foods. 

In addition to application in food estimation, there are methods that have been proposed to estimate the weight of animals using 2D images. Research by \cite{b13} proposed a method to estimate fish weight without contact. \cite{b13} estimated fish weight using the perimeter of the fish. The fish contour information and 3D coordinates of the fish were taken to estimate the perimeter of the fish. After that, perimeter and fish weight information were used to develop a weight estimation model.

\begin{table}[htbp]
\centering
\caption{Information on the Food Dataset}
\begin{tabular}{|c|c|}
\hline
\textbf{Food Type} & \textbf{\textit{Number of Images}} \\
\hline
Cherry Tomato        & 80  \\
Oatmeal              & 50  \\
Steamed Rice         & 40  \\
Stir Fried Spinach   & 122 \\
Sweet Corn           & 252 \\
Grape                & 80  \\
Guava                & 240 \\
Orange               & 86  \\
Papaya               & 160 \\
Pineapple            & 160 \\
Red Apple            & 260 \\
Steamed Bun with Meat& 52  \\
Sweet Potato         & 120 \\
Toast Bread          & 478 \\
\hline
\end{tabular}
\label{tab1}
\end{table}

\section{Proposed Methodology}
It is challenging to divide the dataset for training, validation, and testing because the dataset is an imbalance; features like food weight, container, and orientation must be fairly divided. Therefore, the dataset divides 60\%, 20\%, and 20\% for training, validation, and testing, respectively. 

\begin{figure}[htbp]
\centerline{\includegraphics[width=0.8\linewidth]{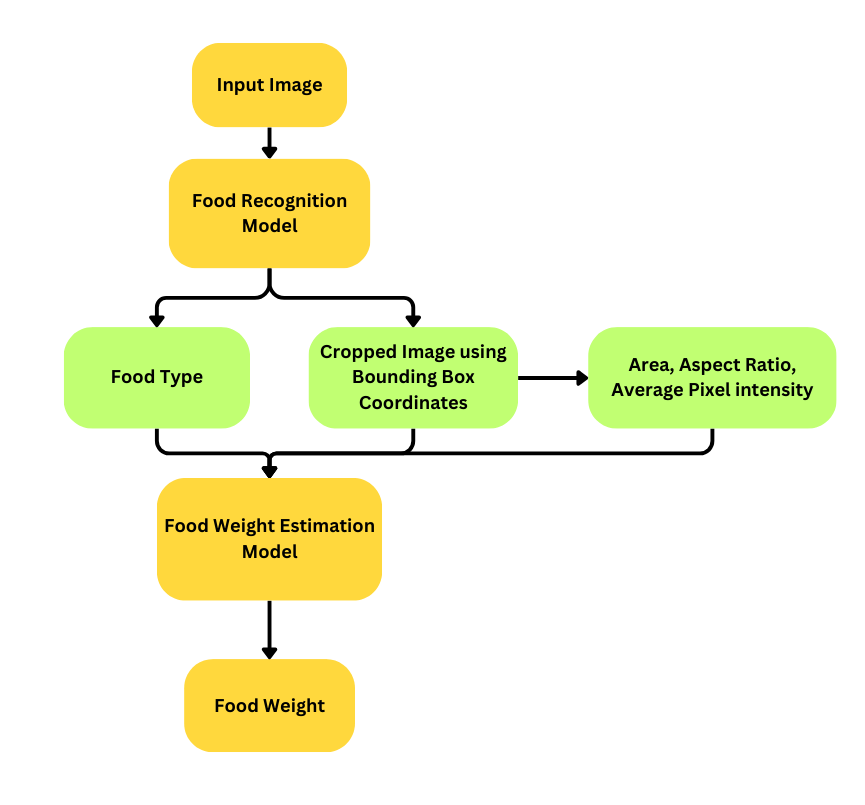}}
\caption{The high-level architecture of the proposed solution}
\label{fig1}
\end{figure}

Fig. \ref{fig1} shows a data flow diagram illustrating the high-level architecture of the proposed solution. The first task is to detect and recognize the food from the image. We use Faster R-CNN \cite{b14} which is a powerful two-stage CNN model for object detection. The two stages are the regional proposal network (RPN), which goes through the input image and proposes candidate regions where objects might be located, and the other stage, which takes proposed regions from the RPN, refines the bounding box coordinates, and predicts the class probabilities for each region. For this task, ResNet \cite{b15} is used as backbone or feature extractor because Resnet is already pretrained on a large dataset, such as ImageNet and it allows them to capture general knowledge about visual patterns. 

Before training, ‘LabelImg’ is used to annotate all the images. After annotating, we resize all the images to (224, 224), which is the standard size for ResNet \cite{b15} and was developed based on computer efficiency, model performance and compatibility with available pretrained models. We have fourteen food types, and the output node is set to fifteen because one node is for the background. The optimizer is used as Adam \cite{b16}, the learning rate is 0.0001, the batch size is 1, and the number of epochs is 10.

\begin{figure}[htbp]
\centerline{\includegraphics[width=0.8\linewidth]{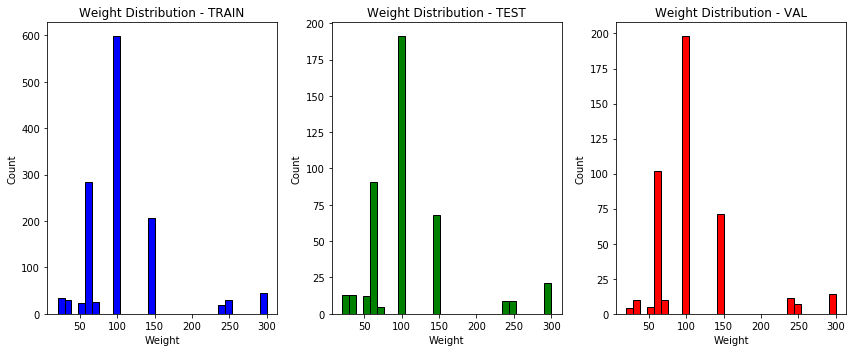}}
\caption{Food weight distribution of training, testing and validation data}
\label{fig2}
\end{figure}

After training Faster RCNN, which can be  used to detect food items with bounding box coordinates and recognize them. All the images are detected, cropped using bounding box coordinates, and also recognized. These data are used to develop the weight estimation model. Fig. \ref{fig2} provides details of the food weight distribution of training, validation, and testing data. 

Features of the weight estimation model are cropped image, food type, image area, aspect ratio, and average pixel intensity. Image area, aspect ratio, and average pixel intensity are calculated using cropped images. Image area means cropped image area. Equation \eqref{eq1} is used to calculate it.

\begin{equation}
Image\: Area = Image\: Height \times Image\: Width\label{eq1}
\end{equation}

Aspect ratio is the proportional relationship between width and height. Equation \eqref{eq2} is used to calculate it. 

\begin{equation}
Aspect\: Ratio = \frac{Image\: Width}{Image\: Height}\label{eq2}
\end{equation}

Average pixel intensity means the average value of all individual pixel intensities. Equation \eqref{eq3} is used to calculate it.

\begin{equation}
Average\: pixel\: intensity = \left ( \frac{1}{N} \right )\sum_{i=1}^{N}p_{i}\label{eq3}
\end{equation}

Where $\mathbf{N}$ means total number of pixels and $\mathbf{p}_{i}$ means intensity of pixel $\mathbf{i}$.

\begin{figure}[htbp]
\centerline{\includegraphics[width=0.8\linewidth]{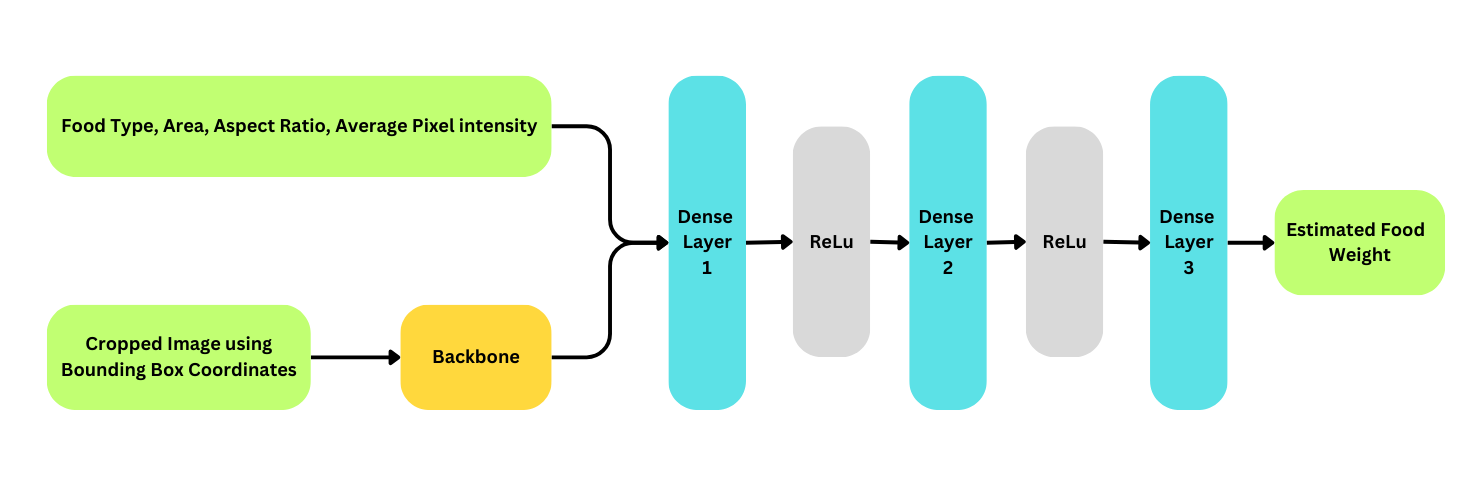}}
\caption{The proposed weight estimation model}
\label{fig3}
\end{figure}

Fig. \ref{fig3} shows the proposed weight estimation model. It takes cropped images and goes through a backbone or feature extractor. MobileNetV3 \cite{b19} is used as the backbone for this task. The output of the backbone is a single output value that concatenates food type, image area, aspect ratio, and average pixel intensity. These five values are used as the input vector of shape (5,1) to the dense layer 1, whose output is a vector of shape (64,1). The output vector from dense layer 1 goes through a Rectified Linear Unit (ReLU) \cite{b17} activation function, which introduces non-linearity into the network by replacing negative values with zero and positive values unchanged. Likewise, the input vector of dense layer 2 is (64,1) and the output vector is (32,1), and the output vector from dense layer 2 goes through a ReLU \cite{b17} activation function. Finally, the dense layer 3 input vector is (32,1) and the output vector is (1,1), which is the predicted food weight value.

We can derive an equation for our weight estimation model. The input X is a vector of size 5 to the dense layer 1.

\begin{equation*}
X = [\alpha, FT, A, AR, API]
\end{equation*}

Where $\alpha$ is the output of the backbone, FT is food type, A is area, AR is aspect ratio, and API is average pixel intensity.

The output vector $\mathbf{Z}_{i}$ for dense layer $i$ is obtained by applying a linear transformation to the input vector $\mathbf{A}_{i-1}$.

\begin{equation}
Z_{i} = W_{i}A_{i-1}+b_{i}
\label{eq4}
\end{equation}

Where $\mathbf{W}_{i}$ is a weight matrix and and $\mathbf{b}_{i}$ is a bias vector for layer $\mathbf{i}$. 

The output vector $\mathbf{A}_{i}$ for ReLU \cite{b17} activation $\mathbf{i}$ is applied element-wise to the elements of $\mathbf{Z}_{i}$.

\begin{equation}
A_{i} = max(0, Z_{i})
\label{eq5}
\end{equation}

$\mathbf{A}_{0}$ is the input vector $\mathbf{i}$. Therefore, the output of the weight estimation model $\mathbf{Y}$ can be written using equations \eqref{eq4} and \eqref{eq5}. 

\begin{equation}
\resizebox{0.9\columnwidth}{!}{$Y=W_{3}\cdot \max(0,(W_{2}\cdot \max(0,(W_{1}[\alpha, FT, A, AR,API]+b_{1}))+b_{2}))+b_{3}$}
\label{eq6}
\end{equation}

Where $\mathbf{W}_{1}$, $\mathbf{W}_{2}$, and $\mathbf{W}_{3}$ are weight matrices of sizes (64, 5), (32, 64) and (1, 32) respectively. $\mathbf{b}_{1}$, $\mathbf{b}_{2}$, and $\mathbf{b}_{3}$ are bias matrices of sizes (64,1), (32,1) and (1,1) respectively.

As preprocessing tasks, after finding the necessary values from the cropped image, all cropped images resize to (224, 224). Then the data augmentation technique, which is the random horizontal flip, is used for training data to increase the diversity of the training dataset. Next, training, validation, and testing data are normalized to the values that are presented in Table~\ref{tab2}. The learning rate, optimizer, and loss function are 0.0001, Adam \cite{b16}, and mean squared error, respectively. For training, batch size and number of epochs are used as 32 and 10, respectively.

\begin{table}[htbp]
\centering
\caption{Data Normalization Mean and Standard Deviation Values}
\begin{tabular}{|c|c|c|}
\hline
\textbf{Data Type} & \textbf{Mean Value ($\mathbf{\mu}$)} & \textbf{Standard Deviation Value ($\mathbf{\sigma}$)}\\
\hline
Training        & 0.4890 & 0.2301\\
\hline
Validation & 0.4844 & 0.2305\\
\hline
Testing & 0.4917 & 0.2287\\
\hline
\end{tabular}
\label{tab2}
\end{table}

\section{Results and Discussions}

The first task is to detect food items with a bounding box and recognize them. Faster R-CNN \cite{b14} and RetinaNet \cite{b18} object detection algorithms were used and selected one according to their performance. The performance of models was evaluated using mAP (mean average precision), including mAP@0.5 and mAP@0.75. mAP, a higher value indicates better performance in object detection. Additionally, we used two metrics: classification accuracy \eqref{eq10} and average Intersection over Union \eqref{eq11}. The results are presented in Table~\ref{tab3}.

\begin{equation}
mAP = \frac{1}{N} \sum_{i=1}^{N} AP_i
\label{eq9}
\end{equation}

Where $\mathbf{AP}_{i}$ is the Average Precision for class $\mathbf{i}$ and $\mathbf{N}$ is the number of classes.

\begin{equation}
\resizebox{0.9\columnwidth}{!}{$Classification\: Accuracy = \frac{Num.\: of\: Correct\: Predictions}{Total \: Number\: of\: Predictions} \times 100$}
\label{eq10}
\end{equation}

\begin{equation}
Average\: IoU = \frac{1}{N} \sum_{i=1}^{N} IoU_i
\label{eq11}
\end{equation}

Where $\mathbf{IoU}_{i}$ is the IoU for the $\mathbf{i}$th prediction and $\mathbf{N}$ is the total number of predictions. 

\begin{table}[htbp]
\caption{Comparison of Food Detection and Recognition Performance using Various Algorithms}
\begin{center}

\resizebox{\columnwidth}{!}{\begin{tabular}{|c|c|c|c|c|c|c|}
\hline
\textbf{Method} & \textbf{Dataset} & \textbf{mAP} & \textbf{$\mathbf{mAP}_{0.5}$} & \textbf{$\mathbf{mAP}_{0.75}$} & \textbf{Classification} & \textbf{Average} \\
& & & & & \textbf{Accuracy} & \textbf{IoU}\\
\hline
\multirow{3}{*}{Faster R-CNN} & Train & 0.8522 & 0.9999 & 0.9999 & 1.0000 & 0.9154 \\
 & Val & 0.8423 & 0.9997 & 0.9667 & 1.0000 & 0.9095 \\
 & Test & 0.8341 & 1.0000 & 1.0000 & 1.0000 & 0.9182 \\
\hline
\multirow{3}{*}{RetinaNet} & Train & 0.7053 & 0.7909 & 0.7878 & 0.8716 & 0.9314 \\
 & Val & 0.7042 & 0.7942 & 0.7912 & 0.8693 & 0.9266 \\
 & Test & 0.7028 & 0.7986 & 0.7892 & 0.8716 & 0.9256 \\
\hline
\end{tabular}}
\label{tab3}
\end{center}
\end{table}

\begin{figure}[htbp]
\centerline{\includegraphics[width=0.8\linewidth]{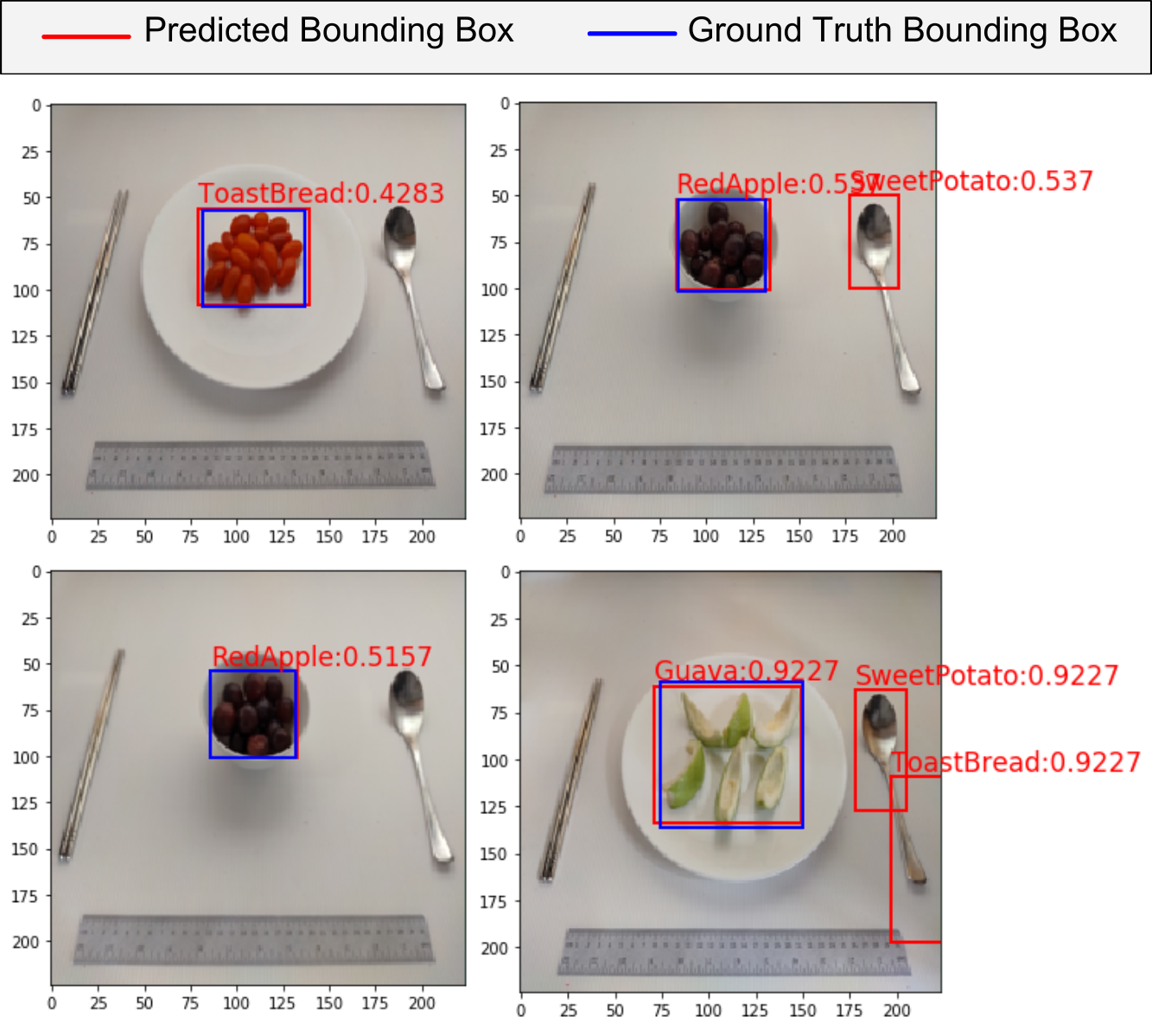}}
\caption{Wrong prediction results of RetinaNet}
\label{fig4}
\end{figure}

Faster R-CNN [14] has a higher mAP in all dataset categories than RetinaNet \cite{b18}. We also considered the mAP at 0.5 and 0.75 Intersection over Union (IoU) thresholds. Faster R-CNN \cite{b14} has higher values than RetinaNet \cite{b18} in every combination. Food type prediction with detection is also very important in this problem because food type is an input to the weight estimation model. Faster R-CNN \cite{b14} classification accuracy is 100\% in all datasets, but RetinaNet \cite{b18} has lower classification accuracy in all datasets. Both algorithms have good value for average IoU. That means the predicted bounding box with high probability score is closely aligned with the actual bounding box. RetinaNet \cite{b18} mAP values are low but have high average IoU values. The reasons might be: the high average IoU might be misleading because it only considers the predicted bounding box with the highest probability score. However, RetinaNet \cite{b18} might still generate other incorrect bounding boxes and misclassify the food name, generating bounding boxes with low-confidence results and also generating a high False Positive rate and a high False Negative rate. Fig. \ref{fig4} shows the low performance of RetinaNet \cite{b18} with examples of low confidence levels, wrong food name predictions, and wrong bounding box predictions of RetinaNet \cite{b18}.

\begin{figure}[htbp]
\centerline{\includegraphics[width=0.8\linewidth]{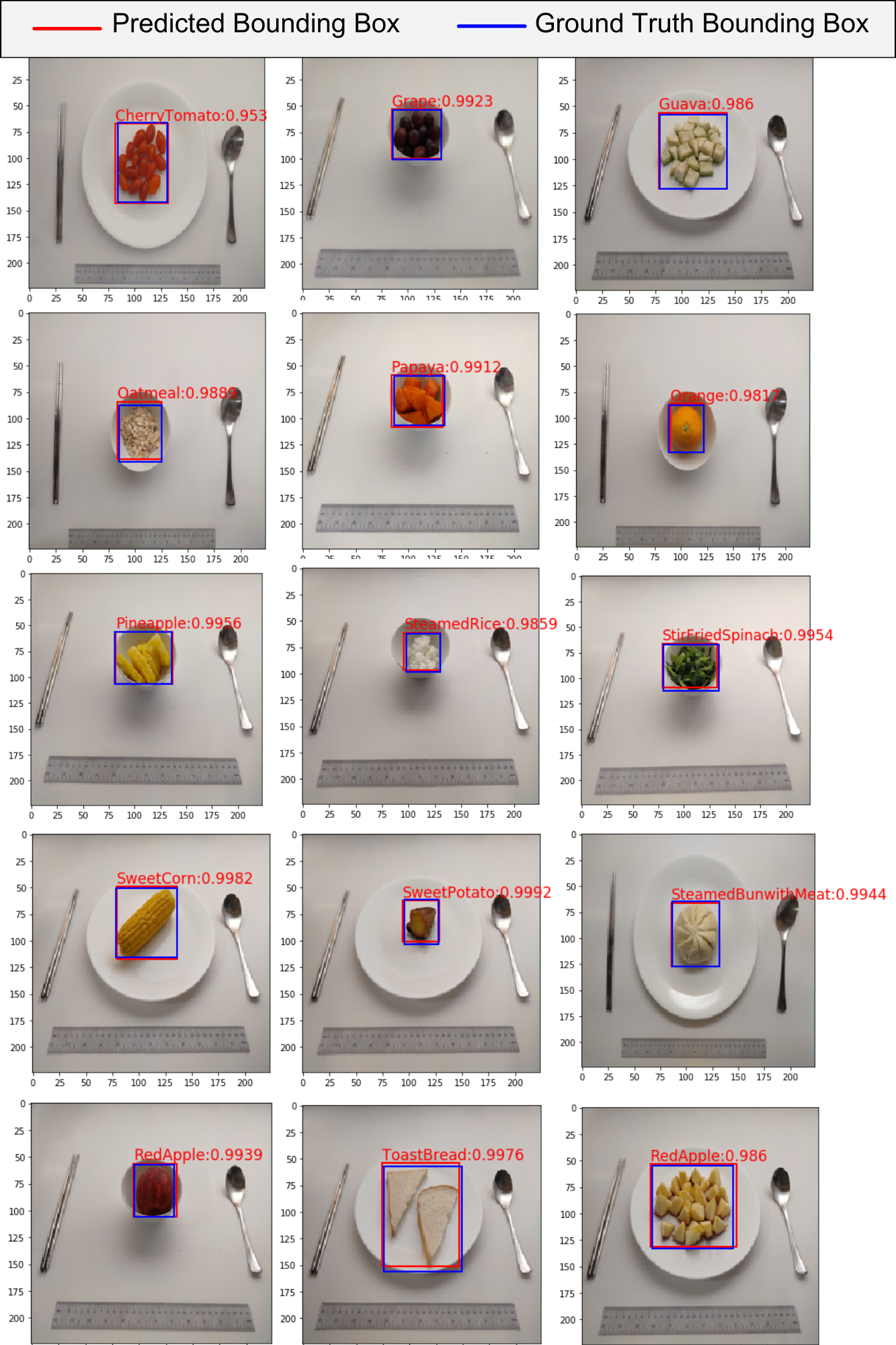}}
\caption{Faster R-CNN Food Detection and Recognition Results}
\label{fig5}
\end{figure}

Therefore, Faster R-CNN \cite{b14} is used for our  food detection and recognition. Fig. \ref{fig5} shows sample images showing the detection and recognition results.  

Next task is to estimate the weight of the food. We have experimented with task performance with different backbones. We used Mean Squared Error (MSE) \eqref{eq12}, Root Mean Squared Error (RMSE) \eqref{eq13}, Mean Absolute Error (MAE) \eqref{eq14}, Mean Absolute Percentage Error (MAPE) \eqref{eq15} and Coefficient of Determination (R-Squared) \eqref{eq16} to evaluate the performance of weight estimation model. 

\begin{equation}
MSE = \frac{1}{n} \sum_{i=1}^{n} (y_i - \hat{y}_i)^2
\label{eq12}
\end{equation}

\begin{equation}
RMSE = \sqrt{\frac{1}{n} \sum_{i=1}^{n} (y_i - \hat{y}_i)^2}
\label{eq13}
\end{equation}

\begin{equation}
MAE = \frac{1}{n} \sum_{i=1}^{n} |y_i - \hat{y}_i|
\label{eq14}
\end{equation}

\begin{equation}
MAPE = \frac{1}{n} \sum_{i=1}^{n} \left|\frac{y_i - \hat{y}_i}{y_i}\right| \times 100
\label{eq15}
\end{equation}

Where $\mathbf{N}$ is the number of samples, $\mathbf{\hat{y}}_{i}$ is the actual value, and $\mathbf{y}_{i}$ is the predicted value.

\begin{equation}
R^2 = 1 - \frac{\sum_{i=1}^{n} (y_i - \hat{y}_i)^2}{\sum_{i=1}^{n} (y_i - \bar{y})^2}
\label{eq16}
\end{equation}

Where $\mathbf{\bar{y}}$ is the mean of the actual values.

MSE (Mean Squared Error) and RMSE (Root Mean Squared Error) are suitable when you want to penalize larger errors more. MAE (Mean Absolute Error) is suitable when  we prefer a metric that is less sensitive to outliers. MAPE (Mean Absolute Percentage Error) is valuable when interpretability is crucial, and you want errors represented as percentages. R-Squared (Coefficient of Determination) helps understand the proportion of variance explained by the model. Using multiple metrics to evaluate your model is a good practice as it provides a more comprehensive understanding of its performance. Each metric may capture different aspects of model performance, and using a variety of them can help you gain insights into how well your model is addressing various aspects of the problem.

\begin{figure}[htbp]
\centerline{\includegraphics[width=0.8\linewidth]{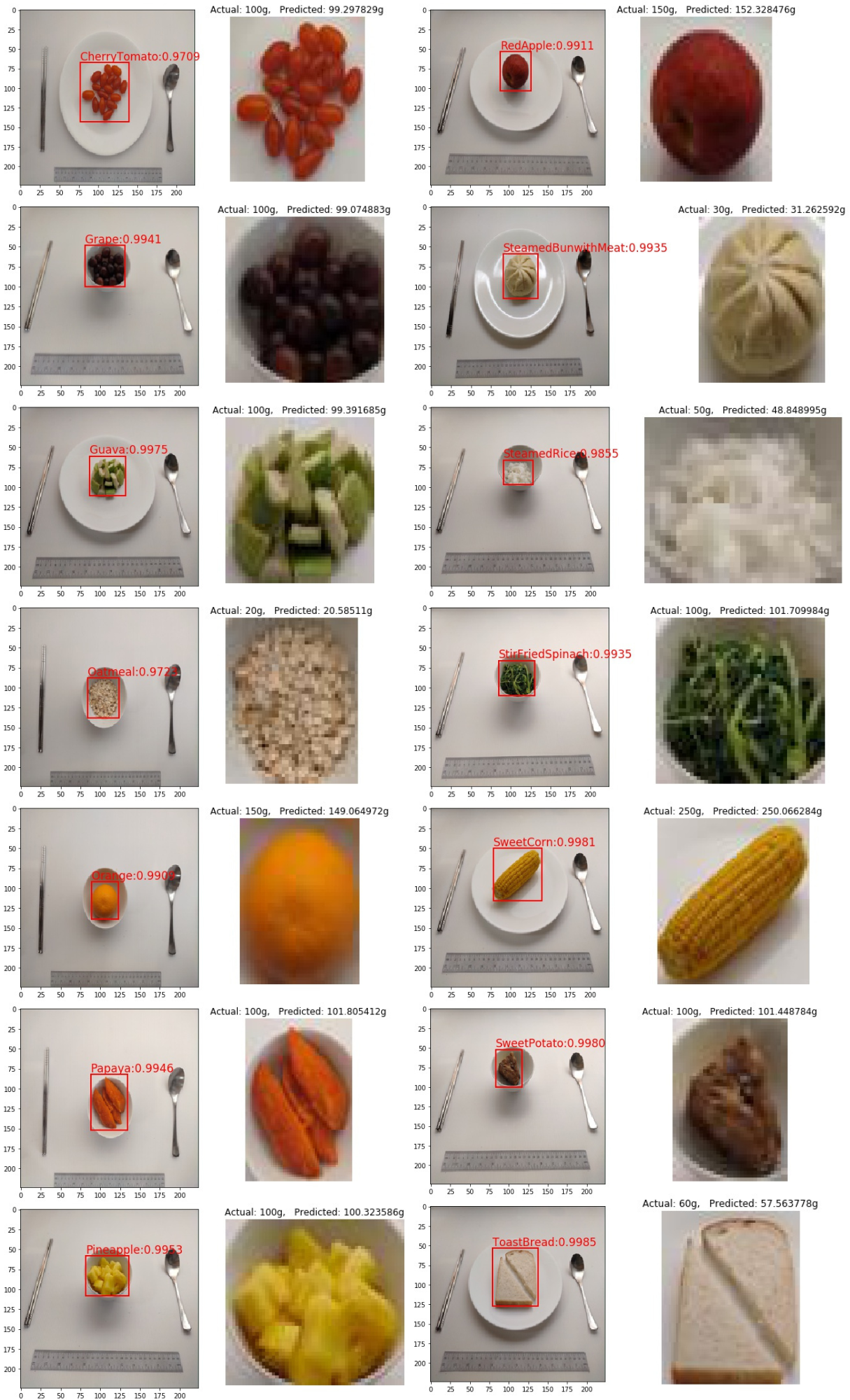}}
\caption{Food Detection and Weight Estimation Results}
\label{fig6}
\end{figure}

We tested five different models, which are MobileNetV3 \cite{b19}, AlexNet \cite{b20}, ResNet50 \cite{b15}, ResNet101 \cite{b15}, and DenseNet121 \cite{b21}. Testing results are the most important result because they represent unseen data for the model. MobileNetV3 \cite{b19} has better values in MSE, RMSE, MAE, and R-Squared. AlexNet \cite{b20} has a better MAPE value. The MAPE value between  MobileNetV3 \cite{b19} and AlexNet \cite{b20} has not a large difference. DenseNet121 \cite{b21} has larger errors when compared with other models. Therefore, MobileNetV3 \cite{b19} was used as the backbone of the weight estimation model. Table~\ref{tab5} presents the average confidence scores, average actual weight, average predicted weight, average weight error, and average absolute weight error of each class, and the final row represents overall. Fig. \ref{fig6} shows the overall system summary. First, it detects the food item with a bounding box, and using a bounding box weight estimation model, it predicts the weight and shows the actual and predicted weight in the figure. 

\begin{table}[htbp]
\caption{Result of the Food Weight Estimation Model}
\begin{center}

\resizebox{\columnwidth}{!}{\begin{tabular}{|c|c|c|c|c|c|c|}
\hline
\textbf{Backbone} & \textbf{Dataset} & \textbf{MSE} & \textbf{RMSE} & \textbf{MAE} & \textbf{MAPE} & \textbf{R-Squared} \\
\hline
\multirow{3}{*}{MobileNetV3} & Train & 23.5711 & 4.8550 & 3.5700 & 0.0502\% & 0.9933 \\
 & Val & 51.7263 & 7.1921 & 4.8986 & 0.0605\% & 0.9848 \\
 & Test & 39.9473 & 6.3204 & 4.8219 & 0.0640\% & 0.9865 \\
\hline
\end{tabular}}
\label{tab4}
\end{center}
\end{table}

\begin{table}[htbp]
\caption{Average Confidence, Actual and Predicted Weights, Weight Error, and Absolute Weight Error for Each Food Class and Total Images}
\begin{center}

\resizebox{\columnwidth}{!}{\begin{tabular}{|c|c|c|c|c|c|}
\hline
\textbf{class} & \textbf{Average} & \textbf{Average} & \textbf{Average} & \textbf{Average} & \textbf{Average} \\
& \textbf{Confidence} & \textbf{Actual} & \textbf{Predicted} & \textbf{Weight} & \textbf{Absolute }\\
& \textbf{Scores} & \textbf{Weight} & \textbf{Weight} & \textbf{Error} & \textbf{Weight Error}\\
\hline
Cherry Tomato & 0.8738 & 100 & 99.5539 & 0.4461 & 0.4461\\
\hline
Grape & 0.9927 & 100 & 100.6670 & -0.6670 & 0.6670\\
\hline
Guava & 0.9920 & 166.6667 & 165.7642 & 0.9024 & 0.9024\\
\hline
Oatmeal & 0.9814 & 20 & 22.25811 & -2.2581 & 2.2581\\
\hline
Orange & 0.9842 & 150 & 151.1659 & -1.1659 & 1.1659\\
\hline
Papaya & 0.9864 & 100 & 99.6441 & 0.3559 & 0.3559\\
\hline
Pineapple & 0.9828 & 100 & 106.3429 & -6.3429 & 6.3429\\
\hline
Red Apple & 0.9675 & 150 & 153.3526 & -3.3526 & 3.3526\\
\hline
Steamed Bun with Meat & 0.9944 & 30 & 31.5970 & -1.5970 & 1.5970\\
\hline
Steamed Rice & 0.9585 & 50 & 53.1813 & -3.1813 & 3.1813\\
\hline
Stir Fried Spinach & 0.9779 & 100 & 102.1309 & -2.1309 & 2.1309\\
\hline
Sweet Corn & 0.9915 & 126.19 & 127.7443 & -1.5543 & 1.5543\\
\hline
Sweet Potato & 0.9968 & 146.6667 & 145.0690 & 1.5977 & 1.5977\\
\hline
Toast Bread & 0.9963 & 60 & 66.5558 & -6.5558 & 6.5558\\
\hline
TOTAL & 0.9769 & 99.9660 & 101.7876 & -1.8217 & 1.8217\\
\hline
\end{tabular}}
\label{tab5}
\end{center}
\end{table}

\section{Conclusion}

This study successfully demonstrates a novel approach for estimating food weight from 2D images using deep learning and computer vision techniques. By utilizing Faster R-CNN for food detection and MobileNetV3 as the backbone for weight estimation, the methodology achieved high accuracy in both detection and weight prediction tasks. The results highlighted significant metrics, including a mean average precision of 83.41\% and an R-squared value of 98.65\%, emphasizing the robustness and reliability of the proposed system. This approach can revolutionize applications in various domains such as healthcare, where accurate dietary assessments are crucial, and in the food industry, where efficient weight estimation can optimize processes and reduce waste. The integration of advanced AI techniques in everyday applications marks a significant step towards leveraging technology for better health and efficiency, paving the way for further innovations in the field.


\end{document}